\documentclass[lettersize,journal]{IEEEtran}
\usepackage{amsmath,amsfonts}
\usepackage{algorithm}
\usepackage{xcolor}
\usepackage{algpseudocode}
\usepackage{array}
\usepackage[caption=false,font=normalsize,labelfont=sf,textfont=sf]{subfig}
\usepackage{textcomp}
\usepackage{stfloats}
\usepackage{url}
\usepackage{verbatim}
\usepackage{graphicx}
\usepackage{multirow}
\usepackage{balance}
\usepackage{tikz}

\def\*#1{\mathbf{#1}}
\usepackage{booktabs}
\usepackage{hyperref}

\newif\ifextremehetero
\begin{document}

% \title{Decentralised Federated Learning over Complex Networks: How to Deal with Extreme Heterogeneity}

% \title{Fully Decentralised Federated Learning over Complex Networks: Overcoming Lack of Coordination and Heterogeneity}
\title{Coordination-free Decentralised Federated Learning on Complex Networks: Overcoming Heterogeneity}

\author{Lorenzo Valerio\IEEEauthorrefmark{1}\IEEEauthorrefmark{2}, Chiara Boldrini\IEEEauthorrefmark{1}\IEEEauthorrefmark{2}, Andrea Passarella\IEEEauthorrefmark{2}, J\'anos Kert\'esz\IEEEauthorrefmark{3}, M\'arton Karsai\IEEEauthorrefmark{3}, Gerardo I\~niguez\IEEEauthorrefmark{4}
\thanks{\IEEEauthorrefmark{1}L. Valerio and C. Boldrini contributed equally to this work.}% <-this % stops a space
\thanks{\IEEEauthorrefmark{2}L. Valerio, C. Boldrini, and A. Passarella are with the Institute of Informatics and Telematics (IIT) of the Italian National Research Council (CNR), Italy. Emails: \{l.valerio,c.boldrini,a.passarella\}@iit.cnr.it.}% <-this % stops a space
\thanks{\IEEEauthorrefmark{3}J. Kert\'esz and M. Karsai are with the Central European University (CEU), Austria. M. Karsai is also with the National Laboratory of Health Security, R\'enyi Institute of Mathematics, Hungary. Emails: kerteszj@ceu.edu, karsaim@ceu.edu}% <-this % stops a space
\thanks{\IEEEauthorrefmark{4}G. I\~niguez is with the Faculty of Information Technology and Communication Sciences of Tampere University (TAU), Finland. Email: gerardo.iniguez@tuni.fi}% <-this % stops a space
% \thanks{Manuscript received April 19, 2021; revised August 16, 2021.}
}

% The paper headers
% \markboth{Journal of \LaTeX\ Class Files,~Vol.~14, No.~8, August~2021}%
% {Shell \MakeLowercase{\textit{et al.}}: A Sample Article Using IEEEtran.cls for IEEE Journals}

% \IEEEpubid{0000--0000/00\$00.00~\copyright~2021 IEEE}
% Remember, if you use this you must call \IEEEpubidadjcol in the second
% column for its text to clear the IEEEpubid mark.

\maketitle

\begin{tikzpicture}[remember picture,overlay]
\node[anchor=south,yshift=10pt] at (current page.south) {\fbox{\parbox{\dimexpr\textwidth-\fboxsep-\fboxrule\relax}{
  \footnotesize{
    This work has been submitted to the IEEE for possible publication. Copyright may be transferred without notice, after which this version may no longer be accessible.
  }
}}};
\end{tikzpicture}

\begin{abstract}
Federated Learning (FL) is a well-known framework for successfully performing a learning task in an edge computing scenario where the devices involved have limited resources and incomplete data representation. The basic assumption of FL is that the devices communicate directly or indirectly with a parameter server that centrally coordinates the whole process, overcoming several challenges associated with it. However, in highly pervasive edge scenarios, the presence of a central controller that oversees the process cannot always be guaranteed, and the interactions (i.e., the connectivity graph) between devices might not be predetermined, resulting in a complex network structure. Moreover, the heterogeneity of data and devices further complicates the learning process. This poses new challenges from a learning standpoint that we address by proposing a communication-efficient Decentralised Federated Learning (DFL) algorithm able to cope with them. Our solution allows devices communicating only with their direct neighbours to train an accurate model, overcoming the heterogeneity induced by data and different training histories. Our results show that the resulting local models generalise better than those trained with competing approaches, and do so in a more communication-efficient way. 
\end{abstract}


% \input{tnnls_body}
%!TEX root = ./tnnls_main.tex
\section{Introduction}
\label{sec:intro}

\IEEEPARstart{W}{e} are currently observing a significant shift within the realm of AI as it moves away from conventional centralized systems and embraces a decentralized approach. This paradigm shift is primarily driven by data generators' changing attitudes at the network's edge\footnote{Following the taxonomy in~\cite{shi2016promise}, user devices such as smartphones are considered edge devices in this work.}. On the one hand, they have become increasingly hesitant to share their private data with third parties, even when promised potential benefits from centralized AI services. On the other hand, the devices located at the network's edge (from now on termed \emph{edge devices}) are becoming ever more computationally capable of performing complex operations on data, including training ML models. 
To address data generators' changing attitudes, decentralized AI systems, such as Federated Learning (FL), have emerged as promising solutions. FL involves keeping the data securely stored on the edge devices, empowering them with a central role in the knowledge extraction process. Rather than sharing their raw data, these devices collaborate by sharing only the parameters (or other related information such as gradients) obtained from their locally trained models. By aggregating these parameters, an enhanced global model is created, which undergoes further refinement through iterative collaboration rounds. This approach ensures privacy of data and enables data generators to actively participate in AI training without compromising sensitive information. 

The standard FL framework assumes a centralized structure where a Parameter Server (PS) oversees the entire process and coordinates the operations of multiple client edge devices (clients for short). The challenges connected to FL are all related to the highly pervasive and heterogeneous environment where the client devices operate. Specifically,  local datasets held by the clients might be non-IID, the client devices might have different and limited capabilities (e.g., computation, memory, battery), and the connectivity might be intermittent, possibly preventing their participation in the distributed training~\cite{Ye:2023aa}. All these challenges can be addressed or mitigated by exploiting the central control of the PS. The PS can synchronize the training by sharing the same model with the clients. It can select which clients' updates to include in the global model~\cite{nishio2019client}, it can mitigate the data heterogeneity during the aggregation phase~\cite{jamali2022federated}, and, more generally, can control the overall progress of the distributed training. However, having a central controller can pose issues such as it being a single point of failure, creating bottlenecks when dealing with millions of devices, and hindering direct collaborations between users. Furthermore, relying on a central controller is not always practical, particularly in highly pervasive edge scenarios. In such cases, devices are expected to operate spontaneously, with minimal or zero prior configuration, without a server. This is especially true in situations where devices are connected via multi-hop networks, allowing communication only with adjacent ones~\cite{lalitha_fully_nodate}.

In light of these considerations, in this paper, we target a system where the central controller is absent. Specifically, our focus is on a highly pervasive environment where numerous devices (e.g., personal mobile devices, IoT devices, etc.) generate data and require an efficient mechanism to collaboratively train a local model without central coordination. We assume that each device can, in general, communicate only with a subset of all devices on a communication graph or network. This graph's topology cannot be predetermined or designed in advance, reflecting the conditions of a realistic edge environment abundant with devices. Hence, it is natural to take and characterize realistic graph topologies as taken from popular models of network science \cite{newman2018networks} on which the decentralized learning process takes place.

In these settings, since the standard FL methodologies based on the central PS are not applicable anymore, we resort to Decentralised Federated Learning (DFL). Briefly, DFL is a generalization of Federated Learning where the devices are connected in a generic graph and typically collaborate only with their immediate neighbours. 
While the absence of a central controller in DFL resolves the single point of failure problem, it introduces new challenges that require investigation. For example, we argue that the information locality, lack of central orchestration, and the network topology may significantly impact the learning process dynamics, specifically how quickly and effectively knowledge embedded in the local models spreads throughout the network.

Topology is not the only aspect that cannot be controlled in a decentralized edge scenario; data distribution and model heterogeneity are additional dimensions to consider. The former is a well-known challenge because, given a specific data-related problem or domain, the data collected by the edge devices may be unevenly represented, resulting in potential under-representation of data on some devices and over-representation on others. This imbalance in data distribution among devices poses a challenge for learning algorithms, which is even more enhanced by the restricted communication between the devices. Model heterogeneity is less explored in the literature and pertains to a unique condition of a decentralized and uncoordinated scenario. Specifically, consider a massive number of devices spontaneously interconnected. Even if all are committed to collaborate to train a common model (i.e., with the same structure) using their local data, it would be unlikely to assume that, at a given point in time, all the devices across the network can agree spontaneously on a common initialization before starting the collaboration. 
Conversely, due to the locality principle, it would be more realistic to consider that each has a model with a different initialization or training status. In a broader sense, this situation reflects a scenario where devices have undergone diverse training and collaboration histories before initiating joint efforts, resulting in an initial condition from the decentralized learning perspective where their models are topologically distant from each other. 

Under these assumptions, the research question we tackle in this paper targets all the three aspects mentioned so far: How to perform an \emph{efficient} decentralized federated learning task in a scenario where (i) there is \emph{no central coordination}, (ii) the interaction between devices is constrained by a \emph{non-trivial topology}, (iii) \emph{data} is distributed heterogeneously across nodes, and (iv) clients' models are \emph{not homogeneously initialized}? 

Our paper answers this question by proposing a decentralized learning algorithm made of two blocks. First, an aggregation function that, considering the differences between the different models induced by the sources of heterogeneity, updates the local models constructively, i.e., without destroying the information learned so far. During the aggregation phase of the local models, the updates are proportional to the relative difference between the local models and the average federated model computed in the local neighbourhood. Second, a knowledge distillation-like approach based on soft-labelling (denoted \emph{virtual teacher}) that we use for improving the local training, overcoming the local lack of information due to data non-IID-ness, and thus obtaining local models with better generalization capability. Our results show that the proposed solution outperforms 
%for the DFL 
DFL and FL competing state-of-the-art approaches in the considered benchmarks. Specifically, the models trained with our decentralized training solution are more accurate than those of the competitors, and the learning scheme is more communication efficient since the devices involved do not exchange any additional information beyond their model parameters. 

The main contributions of this paper are the following:
\begin{itemize}
\item the identification of the challenges connected not only with data heterogeneity but also those related to the lack of coordination between devices involved in a decentralized learning task;
\item a novel and communication-efficient decentralized learning algorithm able to contrast both data heterogeneity and the lack of initial coordination between devices, proving to avoid overfitting in a communication-efficient way;
\item a thorough empirical analysis of the performance of our solution that shows its superior accuracy with respect to the benchmark approaches from the literature. 
\end{itemize}
 
The remainder of the paper is organized as follows. \autoref{sec:relwork} reviews the literature. In \autoref{sec:problem_description}, we define the problem and discuss the approaches that represent the baselines of this work. In \autoref{sec:methodology}, we describe our contribution, and in \autoref{sec:settings} and \autoref{sec:results}, we present the experimental settings and discuss the performance evaluation, respectively. Finally, we draw some conclusions in \autoref{sec:conclusion}. 
%!TEX root = ./tnnls_main.tex
\section{Related work}
\label{sec:relwork}

Extensive research has been conducted on decentralized learning. Initial studies focused on computing the average of decentralized data physically distributed across devices connected through a network topology modelled as a graph. This scenario was particularly relevant for averaging data from sensor networks~\cite{boyd_gossip_2005,aysal_broadcast_2009}. Subsequent investigations delved into Decentralized Gradient Descent (DGD)~\cite{Nedic:2017aa, Yuan:2016aa} and examined its convergence properties under convexity assumptions. In their work~\cite{zeng_nonconvex_2018}, the authors demonstrated that convergence does not strictly require convexity, opening the door for applying DGD to non-convex optimization problems. Apart from DGD, decentralized settings also attracted attention for Stochastic Gradient Descent (SGD). In their publication~\cite{sirb_consensus_2016}, the authors proposed a decentralized SGD algorithm (DSGD) and provided guarantees for its convergence. Furthermore, the work in~\cite{lian_can_2017} analyses the communication efficiency of DSGD, considering that the updates' exchange between nodes is synchronous. Their results show that decentralised SGD can be more efficient than centralised SGD, especially when the communication bandwidth is limited. The asynchronous version of DSGD is analysed in~\cite{Lian2017}. 

Recently, Decentralized Federated Learning, which extends the typical framework of Federated Learning by removing the need for a central parameter server that oversees and coordinates the learning process, has garnered attention from researchers. This new area is gaining significant interest as it combines the privacy advantages of Federated Learning with the potential of decentralized and uncoordinated optimization and learning. 
% Research in this domain considers various factors such as data heterogeneity, non-convex optimization, and the absence of a central server. 
In their work~\cite{Roy2019}, the authors defined a decentralized Federated Learning framework for a medical application involving multiple hospitals collaborating to train a Neural Network model on local and private data. Another approach, proposed in~\cite{Lalitha2019,lalitha_fully_nodate}, adopts a Bayesian-like strategy where the aggregation phase minimizes the Kullback-Leibler divergence between the local model and the received models from peers. While most approaches assume a single local update before sharing parameters or gradients with peers, this assumption is relaxed in~\cite{savazzi_federated_2020} and~\cite{sun_decentralized_2022}. Specifically, \cite{savazzi_federated_2020}~presents a federated consensus algorithm, extending the Federated Averaging model aggregation strategy from FL to decentralized settings, focusing on industrial and IoT applications. \cite{sun_decentralized_2022} proposes Federated Decentralized Average based on SGD, incorporating a momentum term to counteract potential drift introduced by multiple updates and a quantization scheme to enhance communication efficiency. A comprehensive review of decentralised learning algorithms can be found in~\cite{bellavista2021}.

While the works in the literature do consider data-related heterogeneity, i.e., non-IID data across devices, all of them assume to have a homogeneous system from the model point of view, i.e., all the models share the same initialization. To the best of our knowledge, for the first time, we relax this assumption by considering a fully decentralised learning system where not only data is spread in a non-IID fashion in the system but also the models are \emph{initialised heterogeneously}. We propose a solution for dealing with the detrimental effects that such heterogeneity introduces in the learning process.  
%!TEX root = ./tnnls_main.tex
\section{Problem description}
\label{sec:problem_description}

\subsection{Preliminaries}
\label{sec:preliminaries}
 We consider a static graph $\mathcal{G}(\mathcal{V},\mathcal{E})$ where $\mathcal{V}$ denotes the set of vertices (or nodes) and $\mathcal{E}$ the set of edges. We assume that the nodes are user devices and that an edge between a pair of nodes means that the two nodes are willing to collaborate in a supervised learning task (and, thus, can exchange related information, such as models' parameters). Without loss of generality, we focus on a multi-class classification problem. In our system, the graph is a static generic representation of a communication topology between devices. Thus, we assume that the edges (i.e., the links between nodes) can be weighted, i.e., we associate a scalar value $\omega_{ij} \in \mathbb{R}+$ to the edge connecting nodes $i,j \in \mathcal{V}$. In this way, we enable different interpretations of the graph, which can represent, for example, a social network where the link weights represent the level of trust or social intimacy between two nodes. When $\omega_{i,j}=1, \forall i,j \in \mathcal{V}$, the edges represent a simple communication link between two nodes without any other interpretation. 

Each node $i\in \mathcal{V}$ holds a local dataset $D_i$, which contains a set of samples $\{\*x_k,y_k\}_{k=1}^{|D_i|}$, where $\*x_k$ contains the feature values of data item $k$ belonging to local dataset $|D_i|$ and $y_k$ is its label. It holds that $\*x_k \in \mathcal{X}_i$, where $\mathcal{X}_i \subseteq\mathcal{X}$ denotes the local input space and $\mathcal{X}$ denotes the global input space, satisfying $\bigcup_{i=1}^{|\mathcal{V}|}\mathcal{X}_i \subseteq \mathcal{X}$. The $k$-th label $y_k$ for node $i$ is given by $y_k = h(\*x_k;\*w^*)+\eta_i$, where $h(\cdot)$ is a parametric function defined as $h(\cdot;\*w^*):\mathcal{X} \rightarrow \mathcal{Y}$, $\*w^*$ denotes the global ``true parameters'' belonging to the set $W\subseteq\mathcal{W}$ (the parameters space) and $\eta_i$ denotes independent additive noise at node $i$. The global true parameters $\*w^*$ are unknown to the nodes/devices. Each $i$-th device tries to estimate the configuration of the ideal model $h(\cdot;\*w^*)$ through decentralised learning to get an approximated local model, identified by $h(\cdot;\*w_i)$.  Furthermore, we assume that the local training datasets are generated locally on nodes and that there is no data overlapping between them. Moreover, we assume that the distribution of samples per class is \emph{non-IID} across nodes and that the total amount of samples is, in general, different across nodes. This means, following the taxonomy in~\cite{kairouz_advances_2021}, that we focus on quantity and label distribution skew in a cross-device decentralized learning setting.

%================================================================================
\subsection{Partially-decentralised federated learning}
\label{sec:baselines_nondec}
In the following, we briefly introduce the optimization problem associated with standard Federated Learning. While it cannot work in a fully decentralized domain, it serves as a reference point for comparison and aids in understanding how we define the decentralized learning problem tackled in this paper. 
% In the following, we describe the standard baselines for non-decentralized learning. While they cannot work in a fully decentralized domain, they still provide a reference point for comparison to understand the accuracy that could be achieved under a different, more controllable, distributed configuration. For each of them, we briefly summarise the corresponding communication overhead. Communication overhead is an important property to consider in fully decentralized settings, as intense network usage can rapidly drain mobile devices' batteries and consume the mobile data monthly cap depending on the network in use (cellular, WiFi, etc.).
% 
% \noindent {\bf{Central learning:}} All the data are on a central server, and the reference model is trained on it. There is a single model in the network, which we denote with $h(\cdot; \*w_g)$.
% 
% \noindent{}\emph{Communication overhead}: nodes send, once, all their data items to the central server.
% 
% \noindent {\bf{Federated learning - FedAvg~\cite{mcmahan_communication-efficient_2017}:}} FedAvg is the most popular instance of the Federated Learning paradigm. 
% As in all federated learning scenarios, data are scattered across client devices (which would correspond to the nodes in $\mathcal{V}$, which communicate with a central coordinator (PS)). 
% At each communication round $t$, the parameter server sends the current global model to the clients, which trains the model on their local data and then sends it back to the server, which updates the global model with the average of the received models. 
The Federated Learning problem is defined as:
\begin{align}
    % \*w_f^* =& \mathrm{argmin}_{\*w_f} \sum_{i=1}^{|\mathcal{V}|} p_i \frac{1}{|\mathcal{D}_i|}\sum_{j=1}^{|\mathcal{D}_i|}\ell(h(x_j;\*w_f),y_j)
    \*w_f =& \mathrm{argmin}_{\*w} \sum_{i=1}^{|\mathcal{V}|} p_i \underbrace{\frac{1}{|\mathcal{D}_i|}\sum_{k=1}^{|\mathcal{D}_i|}\ell\left(h\left(\*x_k;\*w\right),y_k\right)} \label{eq:fedlearn} \\
    & \phantom{============} F_i(\cdot) \nonumber{} 
\end{align}
where $h(\cdot; \*w)$ denotes the global model, $p_i$ is a scalar representing the relative weight of client $i$ in the learning process, and $\ell$ is the target loss function. Typically~\cite{wang_field_2021}, $p_i$ is set to $\frac{|\mathcal{D}_i|}{\sum_{\forall i \in \mathcal{V}}|\mathcal{D}_i|}$ (which reduces to $\frac{1}{|\mathcal{V}|}$ when local datasets have the same size), effectively weighting clients based on the amount of training data they have. 

The above global optimization problem is solved iteratively, over consecutive communication rounds, in a partially-decentralized way: at each communication round $t=0,\dots, T$, the parameter server sends the current global model to the clients, which train the model on their local data and then send it back to the server, which updates the global model by combining the received models. More technically, client devices perform a certain number of local updates to minimize $F_i(\cdot)$ in \autoref{eq:fedlearn}, at each communication round $t$. The local parameters' set $\*w_i^{(t)} = \textrm{argmin}_{\*w} F_i(\*w)$ resulting from the local optimization (e.g. using SGD) at communication round $t$ on node $i$ is then sent to the parameter server for aggregation. The vanilla aggregation function of FL is Federated Average (FedAvg)~\cite{mcmahan_communication-efficient_2017}, defined as $$\*w_f^{(t)} = \sum_{i=1}^{|\mathcal{V}|} p_i \*w_i^{(t)}.$$ Under common conditions, $\*w_f^{(t)} \sim \*w_f$ for~$t$ sufficiently long~\cite{li_convergence_2020}\footnote{Note, though, that when data are non-IID and the learning rate $\eta$ of SGD is fixed, $\*w_f^t$ will be at least $\Omega(\eta (E-1))$ away from the optimal $\*w_f$, despite convergence. $E$ is the number of epochs for SGD.}. In the most general version of FedAvg, not all clients are involved in the training at each communication round (saving on per-device communications). However, we do not consider this optimization here, for the sake of comparison with the schemes presented next.
% In that case, the aggregation function at the orchestrator becomes $\*w_f^t = \frac{\sum_{i=1}^{|\mathcal{S}^t|} p_i \*w_i^t}{\sum_{i=1}^{|\mathcal{S}^t|} p_i}$, where $\mathcal{S}^t$ denotes the set of clients selected to participate at communication round~$t$.

% \noindent{}\emph{Communication overhead}: The nodes send their models to the central server at each communication round.

% \noindent {\bf{Learning in isolation:}} 
% % There is no central server (neither for the training nor aggregation, like in FedAvg).
% The client devices train their local models $h(\cdot\; ; \*w_i)$ on their local data by minimizing a target loss function $\ell$ as follows:
% %
% \begin{equation}
% \*w_i^* = \mathrm{argmin}_{\*w_i} \sum_{k = 1}^{|\mathcal{D}_i |} \ell(h(\*x_k;\*w_i), y_k)
% \label{eq:isolation}
% \end{equation}
% %
% with $(\*x_k, y_k) \in \mathcal{D}_i$.

% \noindent{}\emph{Communication overhead}: nodes send nothing over the network.

\subsection{Fully-decentralized federated learning}
\label{sec:baselines_dec}

The general learning problem for fully decentralized settings, i.e., with no central server acting as the learning orchestrator, is the same as in Federated Learning (\autoref{eq:fedlearn}). However, this general learning problem cannot be solved anymore by relying on a special node that runs the aggregation and sends back the resulting model to the nodes in the network. While the FL orchestrator is connected to $|\mathcal{V}|$ nodes (effectively mapping the FL network topology to a star graph with the orchestrator at the centre), each node $i$ in DFL is only connected to its neighbours (whose set we denote with $\mathcal{N}_i$). 
% Thus repeated random sampling in $|\mathcal{V}|$ by the orchestrator guarantees that all nodes are picked on average with the same frequency. Hence, all local models will be fairly represented in the global model. Vice versa, whatever nodes in DFL take care of the aggregation can only aggregate the model coming from $\mathcal{N}(i)$.

More in detail, at time $t=0$ the node $i$ trains the model $h(\cdot\; ; \*w_i^{(t)})$ on its data $D_i$ by solving the following problem:
\begin{equation}
\*w_{f,i}^{*(t)} = \mathrm{argmin}_{\*w_i^{(t)}} \frac{1}{|\mathcal{D}_i|} \sum_{k = 1}^{|\mathcal{D}_i |} \ell(h(\*x_k;\*w_i^{(t)}), y_k),
\label{eq:isolation}
\end{equation}
with $(\*x_k, y_k) \in \mathcal{D}_i$. In the following, whenever this does not generate confusion, we simplify the notation by denoting $\*w_{f,i}^{*(t)}$ as $\*w_i^{(t)}$.
% More in detail, at time 0, the model $h(\cdot\; ; \*w_i)$ is, as usual, trained on local data, by minimizing a target loss function: $\*w_i^* = \mathrm{argmin}_{\*w_i} \sum_{k = 1}^{|\mathcal{D}_i |} \ell_i(h(\*x_k;\*w_i), y_k)$. Note that, at least in principle, each node can define its own local loss function for training the local model.
Then, the model $\*w_i^{(0)}$ is sent to all the neighbors in~$\mathcal{N}_i$. 
We assume that nodes entertain a certain number of communication rounds $T$, exchanging and combining local models. At each communication round, a given node $i$ receives the local models from its neighbours in the graph and aggregates them with its local model. The aggregation can be represented as follows:
\begin{equation}
    h(\cdot; \*w_i^{(t)}) \leftarrow \bigoplus_{j \in \mathcal{N}_i} h(\cdot; \*w_j^{(t-1)}).
\end{equation}
Different decentralized aggregation functions will define the operator $\oplus$ differently. Below, we present some state-of-the-art aggregation functions for DFL, against which we will compare our proposal (introduced in~\autoref{sec:contribution}).

\noindent {\bf{Decentralized Federated Average (DecAvg):}}  In the simplest case, at each step $t$, the local model of the given node and the local models from the node's neighbours are averaged FedAvg-style, as specified in the following:
\begin{equation}
    \*w_i^{(t)} \leftarrow \frac{\sum_{j \in \mathcal{N}(i)} \omega_{ij} p_{ij} \*w_j^{(t-1)}}{\sum_{j \in \mathcal{N}(i)} \omega_{ij}},
    \label{eq:decavg}
\end{equation}
%
% where $\omega_{ij}$ are the social weights introduced in Sec.~\ref{sec:preliminaries}, 
where $p_{ij}$ is equal to $\frac{|\mathcal{D}_j|}{\sum_{j \in \mathcal{N}_i} | \mathcal{D}_j|}$ (and captures the relative weight of the local dataset of node $j$ in the neighbourhood of node $i$, similarly to parameter $p_i$ in federated learning's~\autoref{eq:fedlearn} but restricted to the local neighbourhood of node $i$, since everything is decentralized), and $\omega_{ij}$ are the weights in the communication graph $\mathcal{G}$.

This strategy is the natural extension of FedAvg to a decentralized setting: the aggregation is performed not by the central controller (as in federated settings) but by each node, whereby each node averages the model received from a given neighbour based on the strength of the (social) link connecting them.
In this work, we consider two flavours for DevAvg, described below.

\noindent {\bf{DecAvg with initial coordination:}} We assume that nodes can coordinate initially to start learning from a \emph{common model}, i.e., $\*w_i^{(0)} = \*w_j^{(0)}, \forall i,j \in \mathcal{V}$.  This is not generally feasible in realistic settings (unless the network is very small and stable). However, it still provides a good benchmark for what decentralized learning can achieve under ideal conditions.

    % \noindent{}\emph{Communication overhead}: nodes send their models to their neighbours in the graph, and they have to run a coordination algorithm to exchange the initial model parameters from which everybody starts at time zero.

\noindent {\bf{DecAvg with no initial coordination (DecHetero):}} We assume that nodes are \emph{unable} to initially agree on a common model initialization before starting the decentralized training, i.e., they train the same neural network topology (e.g. an MLP) but initialized differently at each node. Thus, $\*w_i^{(0)} \neq \*w_j^{(0)}, \forall i,j \in \mathcal{V}$. This assumption is much more realistic in practical settings but introduces novel challenges that we illustrate in Section~\ref{sec:motivating_example}.

% \noindent{}\emph{Communication overhead}: nodes send their models to their neighbours in the graph.

In all cases, once models are aggregated, the local model is trained again on the local data. % (in this paper, we use learning rate $\eta$ and momentum $\mu$).
As we show next, DecAvg is not suitable for the scenario we consider in this paper, specifically it is not able to cope well with the heterogeneity of data and initial conditions of nodes. Therefore, in the next section, we propose and evaluate a new aggregation strategy able to address the shortcomings we highlight.

%================================================================================

%!TEX root = ./tnnls_main.tex
\section{Methodology}
\label{sec:methodology}

\subsection{Motivating example}
\label{sec:motivating_example}

\begin{figure}
\centering
\includegraphics[scale=.5]{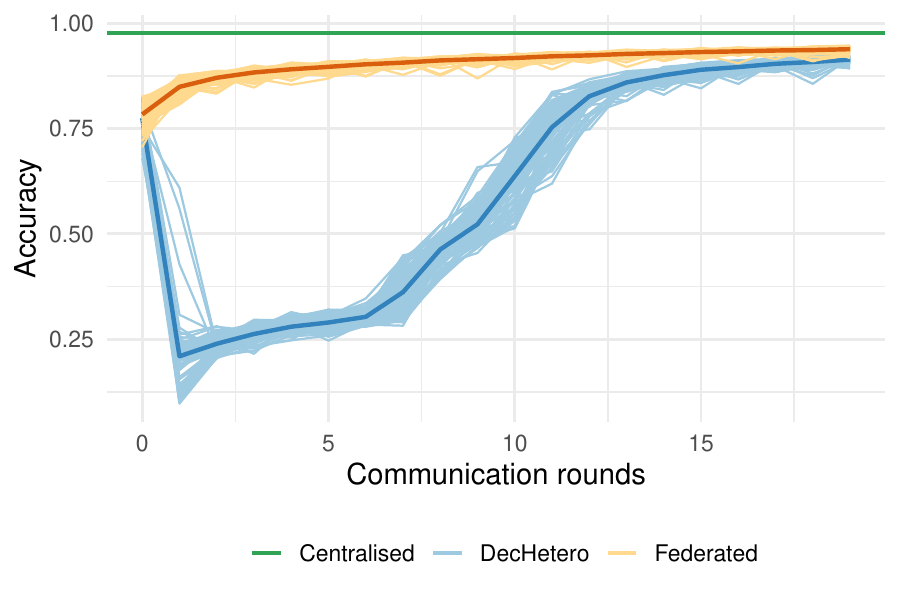}
\caption{Accuracy over communication rounds of centralized, FedAvg and DecAvg without initial coordination (DecHetero). Yellow and light blue curves are for individual devices. Dark blue and orange curves are the averages. An MLP trained by 100 nodes. IID data distribution. Barabasi-Albert connectivity graph.}
\label{fig:motivating}
\end{figure}

DecAvg with no initial coordination (i.e., DecHetero in our terminology) is the most realistic flavor of DecAvg but is expected to suffer from the lack of central coordination in a decentralized environment. In standard FedAvg, the server sends a common initial model to all nodes, which then starts the learning on models sharing the same initial set of parameters. Without the common initialization, local models are expected to associate different features with different groups of parameters (due to the permutation invariance of the hidden layers of neural networks). When this happens, coordinate-wise averaging can be detrimental (because nodes are averaging weights that do not match the corresponding learned features), as exemplified in \autoref{fig:motivating}. The figure reports the accuracy curves obtained using FedAgv (yellow lines) and DecHetero (light blue lines) in a toy scenario where 100 nodes train an MLP on the MNIST dataset\footnote{The MLP configuration is the same as the one used in Section~\ref{sec:settings}.}. The data distribution across nodes is IID, and the underlying connectivity graph is the Barab\'asi-Albert network~\cite{albert2002statistical}, a well-known model with strong degree inhomogeneity that we use here as an example of non-trivial network topology. For the sake of clarity, we also report (i) the average curves for each strategy (dark blue for DecHetero and orange for FedAvg) and (ii) the centralized benchmark (solid green line). Focusing on the DecHetero, at round 0 and after the initial local training, all the nodes reach almost the same accuracy, which dramatically drops after the first aggregation step at round 1. This effect is triggered when the nodes blindly average the received parameters using~\autoref{eq:decavg}. 
The disruption can be interpreted as a synchronization event, after which the nodes' models are topologically more similar. In fact, after round 1, the node's accuracy starts increasing until it almost reaches the performance of the standard federated learning benchmark. 

The drawback of such a simple decentralized approach is that the knowledge contained in the models before the aggregation step has been destroyed almost completely. In a pervasive environment, this might not be acceptable because it would mean that devices lose or waste the results of previous computations each and every time they have the opportunity to collaborate with new nodes. Moreover, it is worth noting that these negative effects are not dependent on the initial data distribution (in this example, data are IID across devices). If we moved towards a non-IID data distribution across devices, these effects would largely magnify, as we will show in \autoref{sec:results}. 
%In this paper, we propose a decentralized training strategy that is able to overcome these sources of heterogeneity, i.e., data distribution and different initialization, as explained in \autoref{sec:contribution}.

% However, the lack of a common initialization can only be solved by forcing the nodes to coordinate in a decentralized way before the learning phase or exploiting strategies that do not require coordination. We believe that an initial coordination round is unsuitable for dynamic decentralized scenarios (percolating the common initialization across the network might be time-consuming, the nodes of the networks may come and go, etc.). For the second approach, we discuss below some possible solutions. 

\subsection{Our contribution}
\label{sec:contribution}
%The results in Section~\ref{sec:motivating_example} have shown the limits of na\"ive averaging in a fully decentralized scenario. Simple coordinate-wise averaging can be detrimental in a decentralized scenario because nodes are averaging weights that do not match the corresponding learned features.
The lack of a common initialization can only be solved by either forcing the nodes to coordinate in a decentralized way before the learning phase or by exploiting strategies that do not explicitly require coordination. As for the latter, solutions exist, e.g., based on Knowledge Distillation. However, they may entail significant computational overhead (as seen in Knowledge Distillation) or communication burdens (for instance, when exchanging gradients alongside models as in~\cite{savazzi_framework_2021}.  On the other hand, in a highly pervasive environment, an initial coordination round is not a feasible precondition. For instance, percolating a common initialization across the network could be time-consuming and even fleeting since the nodes of the networks may come and go. Consequently, while establishing a common initialization might often be impractical, common techniques to avoid the need for the initialization tend to be resource-hungry.
%Moreover, the learning algorithm must be efficient, limiting as much as possible the computation and communication resources required to perform the decentralized training. 

To tackle the above challenge, we base our proposal on two key ideas, the DecDiff aggregation strategy and a virtual teacher mechanism, discussed in the following. Nodes collaborating in a decentralized learning task perform three steps at each communication round: they (i) aggregate the models they received from their neighbours at the previous round into their local model, (ii) retrain the updated aggregate model on the local data, and finally (iii) send the such obtained new model to their neighbours. The strategies we propose affect the aggregation phase and the local training phase. With \emph{DecDiff}, the aggregation function that we propose, we aim to mitigate the issues coming from the lack of coordination in fully decentralized settings. With the \emph{virtual teacher} (VT) mechanism that we leverage for the retraining, we aim to boost the knowledge extraction process from the local dataset without impacting the communication or the computational overhead. 

\subsubsection{Aggregation with DecDiff}
\label{sec:decdiff}

\begin{figure} [!t]
\centering
\includegraphics[width=.8\columnwidth]{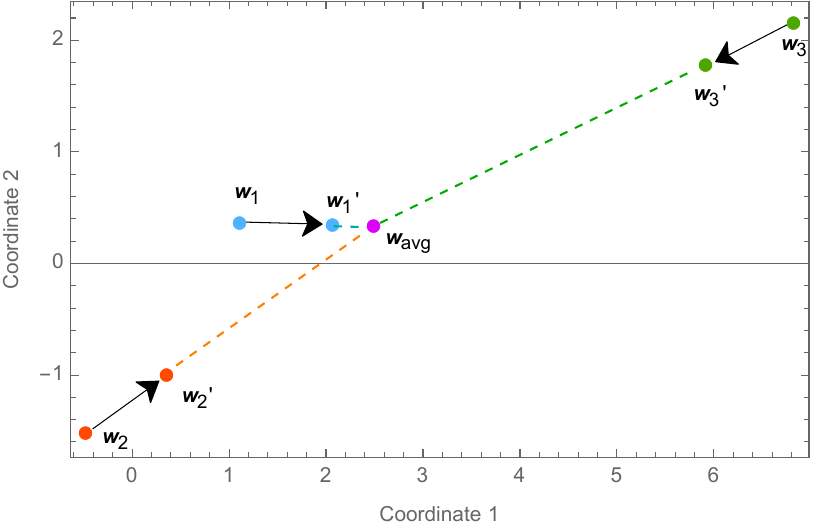}
\caption{Example of parameters' update according to \autoref{eq:decdiffupadte} in a 2D parameters' space.}
\label{fig:decdiff_example}
\end{figure}

The limitations of na\"ive averaging mainly stem from blindly aggregating towards the neighbourhood model's average without first assessing whether this action might cause disruption. To mitigate this problem, our proposal involves updating the local model towards the neighborhood-averaged model, akin to DecAvg, while automatically adjusting the jump inversely proportionally to the $L_2$-distance between the local model and the current neighborhood-averaged model. %   should be less intense when the average model is distant (in $L_2$ norm) from the local one.
\autoref{fig:decdiff_example} reports a pictorial example of our aggregation function. Specifically, the figure shows the initial location of the models of three nodes (i.e., $w_1, w_2, w_3$) and their location after the update (i.e., $w_1', w_2', w_3'$). The length of the arrows depicts the intensity of the update, the dashed lines the resulting distance from the average model. All three models make approximately the same jump, but they start at very different distances from $w_{avg}$. Since $w_1$ is closer to the average model, making a jump is less risky in terms of disruption. Vice versa, $w_3$ is far from the average model, hence it is strategically more convenient to move towards the average but not as much. 
With this approach, during the initial communication rounds where nodes' models are expected to be farther apart, the local model takes small steps towards the average model. With these small steps, the pronounced initial disruption observed in Section~\ref{sec:motivating_example} should be mitigated. Note that the $L_2$-norm implements exactly this reasoning. It penalizes more topologically distant models while accelerating their convergence to a local average model when they start getting closer. This is exactly the rationale of our proposal.
Please note that since in a neural network, the aggregation is done layerwise, it can also be interpreted as an implicit way for implementing a sort of personalization, i.e., different parts of the neural network will embed the outer knowledge at different rates, depending on how different the local representation of reality is with respect to the one in the local neighbourhood. We name this aggregation strategy DecDiff because it effectively factors in the model differences when performing the aggregation.

The above discussion can be translated into the following update rule for DecDiff:
%Formally, at each step $t$, the local model is updated according to the following update rule:
%
\begin{equation}
\*w_i^{(t)} = \*w_i^{(t-1)} + \frac{\bar{\*w}_i^{(t-1)} - \*w_i^{(t-1)}}{|| \bar{\*w}_i^{(t-1)} - \*w_i^{(t-1)} ||_2 + \*s},
\label{eq:decdiffupadte}
\end{equation}
where $1\leq \*s < \infty$ and $\bar{\*w}_i^{(t-1)}$ is the average model.
At each step $t$, we compute the average model $\bar{\*w}_i^{(t-1)}$ (below we explain how in detail) and we compute its distance from the local model $\*w_i^{(t-1)}$ using the $L_2$ distance. Then, the inverse of this distance is used to weigh the parameter update. Since the inverse decreases with increasing distance, the step made toward the average model is smaller when the distance is larger. The constant $\*s$ in the denominator of \autoref{eq:decdiffupadte} prevents an explosion in the magnitude of the updates when the local model starts getting close to the average model in the neighbourhood. In this paper, we set $\*s=1$ because it is the smallest possible value for limiting the influence of the denominator when $0\leq|| \bar{\*w}_i^{(t)} - \*w_i^{(t-1)} ||_2<1$.

The average model $\bar{\*w}_i^{(t-1)}$ is computed as follows:
\begin{equation}
\bar{\*w}_i^{(t-1)} = \frac{\sum_{\forall j \in \mathcal{N}_i} \omega_{ij}p_{ij}\*w_j^{(t-1)}}{\sum_{\forall j \in \mathcal{N}_i}\omega_{ij}p_{ij}}.
\label{eq:decdiff_avg}
\end{equation}
Note that differently from \autoref{eq:decavg}, the average model computed in \autoref{eq:decdiff_avg} by node $i$ does not include the local model. We decided not to include it because the local average model does not substitute the local model (as in DecAvg), but it is used as a reference point to guide the local update. %This means that each node has a slightly different reference average model to follow, depending on the composition of its neighbourhood. 

% \noindent{}\emph{Communication overhead}: nodes exchange their models with their neighbours at each time step.

\subsubsection{Virtual teacher}
\label{sec:virtual_teacher}

While DecDiff affects the aggregation phase of decentralized learning, in the following, we discuss how to boost the training phase without generating additional communication costs. In a situation where nodes only see a small fraction of the whole training dataset, it is important to extract as much knowledge as possible from the data one has. Techniques based on knowledge distillation (KD)~\cite{hinton_distilling_2015} are known to do so. Briefly, in KD, a (typically small) student deep neural network is trained to mimic the outputs of a bigger and well-trained one by minimizing the Kullback-Leibler divergence (KL) between the outputs of the two. Such a system is meant for centralized settings where both models can access the training data. Although there are some attempts~\cite{taya_decentralized_2022} to apply KD in decentralized settings, these works are based on the assumption that a teacher node has enough information to teach a student node how to answer a query correctly. However, in fully decentralized settings, the presence of teacher nodes cannot be guaranteed. Conversely, we assume that all the nodes hold only partial information (since local datasets are non-IID), which might make their models strong only on some portion of data and very weak on others. In summary, we assume that all the nodes' models are locally weak from the Teacher-Student (TS) paradigm point of view. 
However, according to \cite{yuan_revisiting_2020}, not only is it still possible to extract knowledge from poorly trained models through knowledge distillation, but in such a case, we can also overcome the need for a real teacher and use a virtual one implemented as soft labels. Specifically, each node defines a soft-labelling scheme for the local data and uses it to train the local model under the KD paradigm. So, rather than relying on a trained teacher model, we can ``emulate'' one. 
To this aim, the probability $p_t(y)$ that the virtual teacher assigns label $y$ to a data of class $c$ is as follows:
\begin{equation}
    p_t(y) =    \begin{cases}
                \beta & y=c\\
                (1-\beta)/(|\mathcal{L}|-1) & y \neq c,
                \end{cases}
    \label{eq:vt}
\end{equation}
where $c$ is the correct label, $|\mathcal{L}|$ is the total number of classes, and $\beta$ is the probability yielded for data belonging to that class. For a good teacher, it is reasonable to assume $\beta \geq 0.9$. The excess to 1 is equally spread among all other labels. Thus, during the training phase, we minimize the Kullback-Liebler divergence between the (hand-crafted) local virtual teacher and the output of the model:
%
% \begin{multline}
%     \mathrm{argmin}_{\*w} \sum_{j \in \mathcal{D}_i} (1-\alpha) \ell\left(y_j, h_i(\*x_j; \*w)\right) + \\
%     \alpha \ell_{KL}\left(p_i\left(\*x_j; \*w\right),p_{t}(l_j)\right),
%     \label{eq:decvt}
% \end{multline}
\begin{equation}
    \mathrm{argmin}_{\*w} \sum_{j \in \mathcal{D}_i} \ell_{KL}\left(p_i\left(\*x^i_j; \*w_i^{(t)}\right),p_{t}(y^i_j)\right),
    \label{eq:decvt}
\end{equation}
where $p_i\left(\*x^i_j; \*w_i^{(t)}\right)$ are the output probabilities extracted from $h_i(\*x^i_j; \*w_i^{(t)})$ and $p_{t}(y^i_j)$ are those returned from the virtual teacher model in Equation~\ref{eq:vt}. From a computation point of view, such a method results in a simple soft-labelling of the local dataset according to the $\beta$ parameter, and it does not require running any other external model to obtain the outputs for computing the KL-loss. 

\subsection{Proposed Algorithm}
The resulting algorithm is described in Algorithm \ref{alg:ddvt}. 
All the $|\mathcal{V}|$ nodes in the system start by initializing the local model independently such that given two nodes $i,j$, $\*w_i \neq \*w_j, \forall i,j\in \mathcal{V}$ with $i\neq j$ and, by setting the parameter $\beta$ in \autoref{eq:vt} for creating the local soft labels. The local training is performed according to the standard procedure. Each node $i$ trains the local model for a certain number of local epochs $E$ (which is not necessarily the same at all nodes), minimizing \autoref{eq:decvt} (lines 3-8). In this paper, we use SGD with momentum, but the procedure holds for any optimization algorithm. Once the local training is complete, a node $i$ sends its local model to the other nodes in its neighbourhood ($\mathcal{N}_i$) and receives their models in return (lines 11-12). Note that we do not impose any synchronization in this phase; thus, a node might receive a model from all or just a fraction of its neighbours at each communication round. 
The set of models $\{\*w_j\}_{j\in\mathcal{N}_i}$ collected by node $i$  from its neighbourhood are used to compute the local average model according to \autoref{eq:decdiff_avg} (line 13) and perform the aggregation step according to \autoref{eq:decdiffupadte} (line 14).
% $\bar{\*w}_i$ using a simple weighted average
% \begin{equation}
%     \bar{\*w}_k = \sum_{n\in\mathcal{N}_k} \rho_{k,n}\*w_n
%     \label{eq:loc_avg}
% \end{equation}
% where the weighting parameter $\rho_{k,n}=\frac{|\mathcal{D}_n|}{\sum_{n\in\mathcal{N}_k} |\mathcal{D}_n|}$ is the portion of data that each node in the neighbourhood has (line 13). This weighting parameter is similar to the one initially proposed in \cite{mcmahan_communication-efficient_2017}, apart from the fact that it does not include the node $k$. In this way, we build a reference average model that node $k$ uses \emph{as a driver} for performing the aggregation step (line 14). 
The nodes in the system perform the DecDiff+VT algorithm for $R$  communication rounds. 
\begin{algorithm}
\caption{DecDiff+VT Algorithm}\label{alg:ddvt}
\begin{algorithmic}[1]
% \Require $n \geq 0$
% \Ensure $y = x^n$
% \State $y \gets 1$
% \State $X \gets x$
% \State $N \gets n$
% \While{$N \neq 0$}
% \If{$N$ is even}
%     \State $X \gets X \times X$
%     \State $N \gets \frac{N}{2}$  \Comment{This is a comment}
% \ElsIf{$N$ is odd}
%     \State $y \gets y \times X$
%     \State $N \gets N - 1$
% \EndIf
% \EndWhile
% \State All $i\in|\mathcal{V}|$ nodes initialise the local model $\*w_i$ at random. 
\State Randomly init of $\*w_i, \forall i \in \mathcal{V}$ \Comment{Local model initialization}
\State Set the number of local training epochs $E$, the batch size $bs$ and $B=E*bs$ the total number of minibatches sampled during the local training phase between two communication rounds
\ForAll{$t=0 \dots T$ all nodes $i\in \mathcal{V}$ in parallel}
% \For{e=0 \dots E}\Comment{Local training}
\For{$b=0 \dots B$} \Comment{Local training}
\State Sample the $b$-th mini-batch $\boldsymbol\xi_b=(\*X_b,\*Y_b)$
\State Obtain soft labels $\tilde{\*Y}_b$ as in \autoref{eq:vt} \Comment{output: probability vector}
\State Compute train loss defined in \autoref{eq:decvt}
\State Update model $\*w_i = \mathrm{SGD}(\eta,\mu)$
\EndFor
% \EndFor
\State Send $\*w_i$ to the nodes in $\mathcal{N}_i$
\State Receive models from neighbours $\{\*w_j\}_{j\in\mathcal{N}_i}$
\State Compute local average model as in \autoref{eq:decdiff_avg}
\State Update local model as in \autoref{eq:decdiffupadte}
\EndFor
% \EndFor
\end{algorithmic}
\end{algorithm}

\section{Experimental settings}
\label{sec:settings}

In order to evaluate the learning strategy proposed in Section~\ref{sec:contribution}, we focus on a fully decentralized collaborative image classification task. For the experiments, we use a custom Python simulator (\textsc{SaiSim}) that integrates graph generation (via the \texttt{networkx} library), deep learning functionalities (via the \texttt{PyTorch} library) and offers novel routines to support fully decentralized learning\footnote{A link to the simulator code will be released in the camera ready of the paper if accepted.}. We perform four replicas for each scenario described below, reporting $95\%$ confidence intervals when relevant. The same simulator has also been used in our prior work~\cite{palmieri2023effect,palmieri2023exploring}.

\subsubsection{Social network topology}
\label{sec:topology}

\begin{figure}[t]
\centering
\includegraphics[scale=0.8]{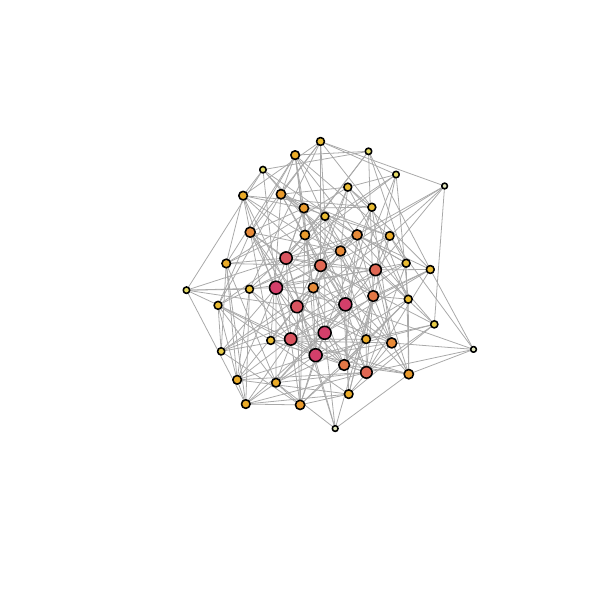}
\caption{The Erd\H{o}s–R\'enyi graph we used for simulations. The vertex size and colour (from red to yellow) are proportional to the node degree.}
\label{fig:graph}
\end{figure}

We consider an Erd\H{o}s–R\'enyi graph with 50 nodes and edge creation probability set to $0.2$, which is well above the critical value $0.078$ (obtained as $\frac{\ln(50)}{50}$ \cite{erdHos1960evolution}). This means that the network is almost surely connected. The latter property is important to avoid degenerate situations where nodes exist that cannot communicate with anyone else, which would not be suitable for assessing the properties of decentralized learning. We believe that investigating the impact of network connectivity on the performance and robustness of decentralized learning is a critical research problem to be tackled, but it does not fit within the scope of this paper, and we leave it as future work. Preliminary investigations of this topic have been presented in ~\cite{palmieri2023effect,palmieri2023exploring}. The specific instance of the Erd\H{o}s–R\'enyi graph used for our simulation is shown in~\autoref{fig:graph}.

\subsubsection{Datasets}
\label{sec:datasets}

We consider the following different datasets.

\noindent MNIST: This dataset is a collection of handwritten digits. It consists of 60,000 training examples and 10,000 test examples. Each example is a grayscale image of 28x28 pixels, representing a single digit from 0 to 9. The digits are handwritten by various individuals, making the dataset diverse in writing styles and quality.

\noindent FASHION-MNIST: This dataset contains images of ten different fashion item categories. The number of training/test examples is the same as for MNIST, as well as the grayscale image pixel size.

\noindent EMNIST: This dataset is an extension of the original MNIST dataset, including handwritten digits and handwritten letters. Here, we consider only the split EMNIST Letters, which focuses exclusively on uppercase letters. It comprises 26 classes, with 20,800 training examples and 3,280 test examples.

\subsubsection{Data allocation}
\label{sec:data_allocation}

\begin{figure}[t]
\centering
\includegraphics[width=.95\columnwidth]{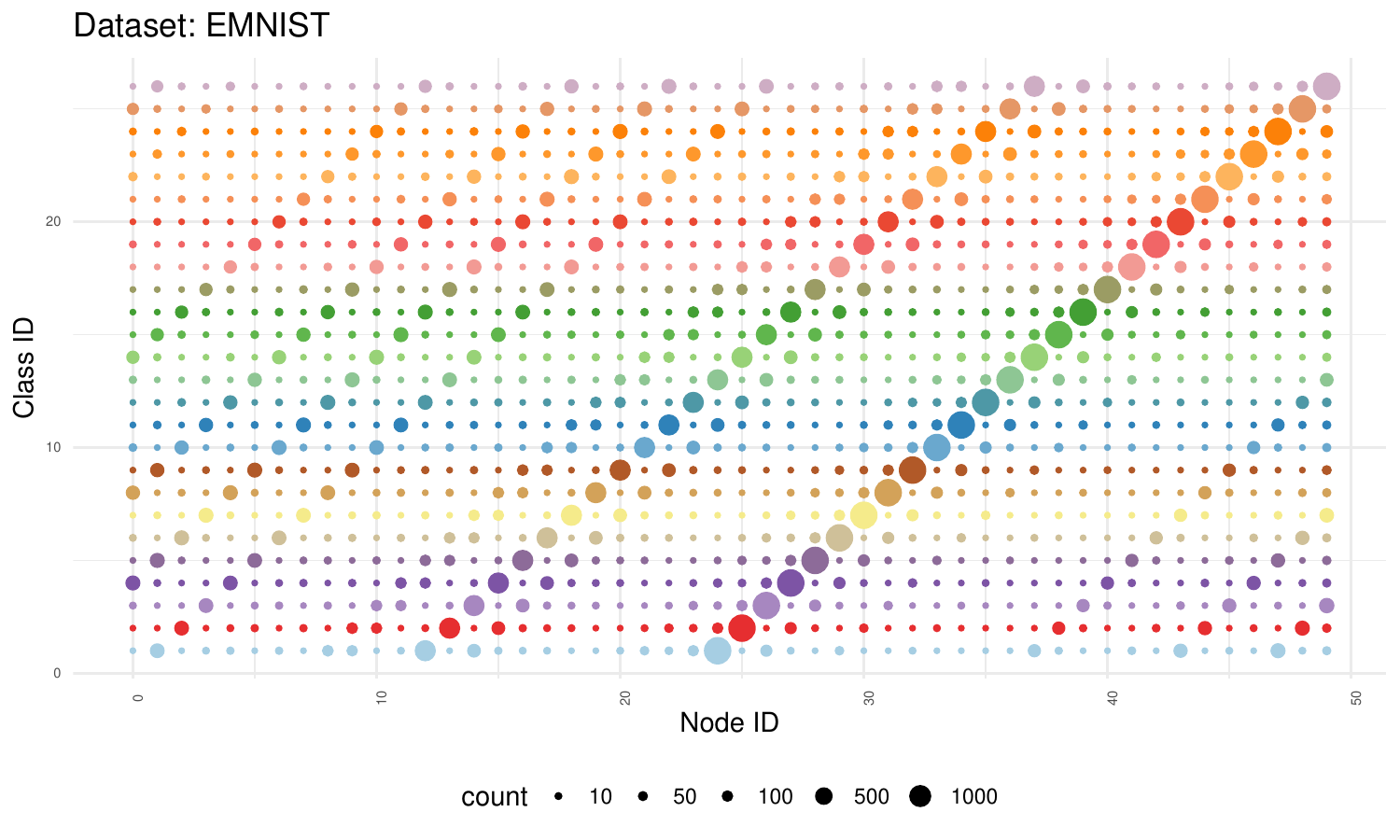}
\caption{Example of data distribution over nodes for EMNIST.}
\label{fig:data_distribution}
\end{figure}

In order to reproduce a realistic scenario, data items from the training set are distributed in a non-IID way across nodes. Specifically, class images are assigned to nodes according to a Truncated Zipf distribution with exponent $\alpha_{zipf}=1.26$ to guarantee sufficiently skewed distributions, as explained next. We draw 50 Zipf samples for each class whose values correspond (approximately, see the limitation of boundary effects below) to the number of images for that class assigned to the node. 
This approach results in a highly skewed data distribution, with one node containing the majority of images for a particular class.  In order to avoid boundary effects caused by nodes with zero images for a certain class, we guarantee that all nodes see at least some images for each class, however few. An example allocation for the EMNIST dataset is shown in \autoref{fig:data_distribution}. Such skewed data distribution implies that no node alone can perform well on the test set where all classes are present. Consequently, collaboration among nodes becomes crucial.

Due to the potentially disruptive effect on the learning process of skewed data distribution across nodes, it is important to capture quantitatively how skewed the data allocation is in each simulation run. To this aim, we leverage the Gini Index (GI), which is a standard inequality metric. The Gini Index ranges from 0 (perfect equality) to 1 (absolute inequality). Our goal is to tackle a fairly heterogeneous data distribution. Thus, we work in the $[0.7, 0.85]$ range. For all plots concerning single simulation runs, we report the Gini index associated with the instance of data heterogeneity in that run. 
\ifextremehetero
To complement this analysis, we consider an extremely skewed distribution ($GI > 0.9$) in Section~\ref{sec:extreme_heterogeneity}.
\fi

\subsubsection{Local model architecture}
\label{sec:local_model_architecture}

\begin{table}
\centering
\caption{NN models architectures used in this paper}
\label{tab:nn_topology}
\begin{tabular}{llp{.1\textwidth}ll}
\toprule
\textbf{Dataset} & \textbf{Model} & \textbf{Structure} & \textbf{Activation} & \textbf{Kernel size} \\
\midrule
MNIST & MLP & FC:512,256,128  & ReLU & - \\
Fashion & CNN & Conv2d:32,64\newline FC:9216,128 & ReLU & 3$\times$3 \\
EMNIST & CNN & Conv2d:32,64\newline MaxPool(2)\newline Dropout(.25) \newline FC:9216,128\newline Dropout(.5)\newline FC:128 & ReLU & 3$\times$3 \\
\bottomrule
\end{tabular}
\end{table}

We use different local model architectures for the different datasets, as reported in \autoref{tab:nn_topology}. As an optimizer, we use SGD with the following configurations. For MNIST, we set the learning rate $\eta = 0.001$ and the momentum $\mu = 0.5$. For FASHION-MNIST and EMNIST, we set the learning rate $\eta = 0.001$ and the momentum $\mu = 0.9$.

For all strategies except Centralized and FedAvg, the model parameters are drawn at random locally at each node. This implies that nodes start with different model initialization, as expected in realistic decentralized settings.

For all the strategies except DecDiff+VT, the function to optimize during the local training is the cross-entropy loss function $$\mathcal{L}_{CE}=-\frac{1}{B}\sum_{k=1}^B\sum_{c=1}^C y_{k,c} \log(p_{k,c})$$ with mean reduction, where $B$ is the batch size, $C$ is the number of classes, $y_{k,c}$ is the $c$-th of the one-hot encoding of label $y_k$ and $p_{k,c}$ is the Softmax output of the neural network model $h(\cdot,\*w)$ over the sample $x_k$.

\subsubsection{Benchmarks}
\label{sec:benchmarks}

We compare decentralized learning against the following benchmarks.

% \item[DecTS-Oracle] The upper bound to teacher-based learning is when the teacher is as good as they get, i.e., the teacher model is exactly the centralized model obtained with \emph{Central learning}. This baseline is not meaningful in practice, but it provides an estimate of the potential of TS learning: if DecTS-Oracle does not beat another target strategy, there is no designing an alternative teacher that can be that [@Lorenzo: actually, this might not be true? Some noise could prove useful?]. 
\noindent {\bf{Centralized:}} All the data are on a central server, and the reference model is trained on it. There is a single model in the network, which we denote with $h(\cdot; \*w_g)$.

\noindent {\bf{Isolation (ISOL):}} 
% There is no central server (neither for the training nor for the aggregation, like in FedAvg).
The client devices train their local models $h(\cdot\; ; \*w_i)$ on their local data by minimizing a target loss function $\ell$.

\noindent{\bf{DecHetero}:} The standard Decentralised Federated Learning without initial coordination as described in \autoref{sec:baselines_dec}. 

\noindent {\bf{CFA:}} Consensus-based Federated Averaging has been proposed by~\cite{savazzi_federated_2020}. Aggregation is performed as follows:
\begin{equation}
    \*w_i^{(t)} = \*w_i^{(t-1)} + \epsilon \sum_{j \in \mathcal{N}_i} p_{ij} \left( \*w_j^{(t-1)} - \*w_i^{(t-1)} \right),
    \label{eq:cfa}
\end{equation}
with $p_{ij} = \frac{|\mathcal{D}_j|}{\sum_{k \in \mathcal{N}_i} |\mathcal{D}_i|}$ and $\epsilon \in \left(0, \frac{1}{\Delta}\right)$, with $\Delta=|\mathcal{N}(i)|$. As in the follow-up work~\cite{savazzi_framework_2021}, we set $\epsilon = \frac{1}{\Delta}$. With respect to our approach, in CFA the local model is updated based on the individual models of neighbours (instead of their average), thus without taking into account an overall reference point defined at the level of the neighborhood.
% but in~\cite{savazzi_federated_2020} they seem to assume $\Delta = 1$ (with $\epsilon_t$ set to 0.5 or 1)
% When $\epsilon_t = 1$, CFA is equivalent to DecAvg, while when $\epsilon_t = |\mathcal{N}_i|$, the local model weighs much more than the others (i.e., $\frac{1}{\mathcal{N}_i}\left[(\mathcal{N}_i-1)\*w_i + \sum_{k \in \mathcal{N}_i \bigcup \{i\}} \alpha_{i,k} \*w_k\right]$).

\noindent {\bf{CFA-GE:}} It has been proposed in~\cite{savazzi_federated_2020}, as a variation of CFA, where \emph{both} models and local gradients are exchanged at each communication round. Specifically, after the aggregation (performed as in~\autoref{eq:cfa}), node $i$ sends the model parameters $\*w_i^{(t)}$ to its neighbours and asks them to use the model to compute the gradients on the neighbours' local data. Once computed, the neighbours return the gradients to node $i$, which then updates model  $\*w_i^{(t)}$ with the received gradients. In our implementation, we use the speed-up implementation proposed in~\cite{savazzi_federated_2020}.
This model is expected to perform well due to the exploitation of gradients trained on neighbours' data. However, it has a high communication and computation overhead since nodes are expected to exchange their gradients and models with their neighbours and train their neighbours' models on local data. While this could be infeasible in practice on user devices, it is still interesting to consider this approach as one of the benchmarks. We anticipate that in our proposal a similar effect, in terms of accuracy, is obtained through the use of the virtual teacher, with the advantage of a significant cut in the communication overhead.

\section{Results}
\label{sec:results}

\subsection{Performance comparison with non-IID data}
\label{sec:accuracy}

\begin{table*}[ht]
\centering
\caption{Average accuracy on MNIST, Fashion-MNIST, EMNIST, with 95\% confidence interval}
\label{tab:avg_acc_all}
\begin{tabular}{@{}lllll@{}}
\toprule
& & MNIST & Fashion-MNIST & EMNIST \\
& Method & \bf{Avg accuracy} & \bf{Avg accuracy} & \bf{Avg accuracy}\\ 
\midrule
\multirow{2}{*}{\par{Standalone baselines}}&Centralised & 0.9824 $\pm$ 0.0000 & 0.9181 $\pm$ 0.0000 & 0.9027 $\pm$ 0.0000 \\ 
& ISOL & 0.6473 $\pm$ 0.0900 & 0.7594 $\pm$ 0.0392 & 0.4654 $\pm$ 0.0864 \\ 
\midrule
Partially decentralised baseline & FED & 0.9410 $\pm$ 0.0156 & 0.8939 $\pm$ 0.0034 & 0.8344 $\pm$ 0.0211 \\ 
\midrule
\multirow{3}{*}{Decentralised SOTA}& DecHetero & 0.9071 $\pm$ 0.0249 & 0.8837 $\pm$ 0.0029 & 0.8301 $\pm$ 0.0278 \\ 
& CFA & 0.8975 $\pm$ 0.0306 & 0.8568 $\pm$ 0.0048 & 0.7807 $\pm$ 0.0345 \\ 
& CFA-GE & 0.9460 $\pm$ 0.0048 & 0.8564 $\pm$ 0.0130 & 0.8607 $\pm$ 0.0092 \\ 
\midrule
% \multirow{2}{*}{Decentralised Proposal}& DecDiff & 0.9373 & 0.01526 & 0.8838 & 0.00459 & 0.8214 & 0.02575 \\ 
Decentralised Proposal & DecDiff+VT & \textbf{0.9530} $\pm$ 0.0154 & \textbf{0.8904} $\pm$ 0.0043 & \textbf{0.8653} $\pm$ 0.0128 \\ 
\bottomrule
\end{tabular}
\end{table*}
%Whatever the specific decentralized learning strategy, the goal of decentralized learning is to approach the performance of its centralized counterpart. More realistically, given that the latter is an ambitious goal for decentralized learning, we strive to perform similarly to a centrally controlled approach as federated learning. Therefore, we compare the results of our proposed solution against three different types of benchmarks: (i) the standalone baselines where no cooperation between devices is involved, (ii) the standard federated learning framework where a parameter server coordinates the learning process and (iii) the decentralized framework where there is cooperation but no coordination between devices.
%In the following, we will refer to \autoref{tab:avg_acc_all}.
\autoref{tab:avg_acc_all} shows the results we discuss in this section, where we focus on the accuracy achieved by the different learning strategies. Specifically, we report the average accuracy across all the nodes in the network at the last communication round. For Centralized, the value reported in the table is not an average, since Centralized only involves a single central server.  

\subsubsection{Comparison against standalone and partially decentralised baselines}the performance of the standalone approaches, i.e., Centralised and Isolation, constitutes the two extremes of the obtainable performance range. Given a DNN model and a dataset, the former represents the upper bound that all the (fully or partially) decentralized learning approaches could achieve. In contrast, the latter represents the lower bound we would obtain without cooperation between devices. The gap between the two expresses the maximum gain from using a decentralized learning scheme. In our settings and depending on the specific learning configuration, the average maximum possible gains (i.e., the average margin between the accuracy of Isolation and Centralised) are 34 (MNIST), 16 (Fashion-MNIST) and 44 (EMNIST) percentage points. Finally, the Federated Learning (FL) performance stands between Centralized and Isolation. It represents a second reference benchmark since the operating conditions are more favourable w.r.t. a fully decentralized system, i.e., the parameter server is a single aggregation point, and all the clients train the same global model. Our solution, DecDiff+VT, achieves comparable or even better performance than FL with FedAvg across all the datasets. Precisely, on Fashion-MNIST, the final average accuracy is statistically equivalent to FL, while on MNIST and EMNIST nodes, predictions are 1.20\% and 3.09\% more accurate. When compared against Isolation, the accuracy of DecDiff+VT improves by 30.57 (MNIST), 13.1 (Fashion-MNIST), and 39.99 (EMNIST) percentage points, hence approaching the maximum gains discussed above and providing a good approximation of Centralised despite the challenging settings.

\subsubsection{Comparison against simple decentralized baselines}the above results suggest that our solution can overcome difficulties induced by two sources of heterogeneity: data and models' initialization. This is also confirmed by looking at the performance of the decentralized benchmarks. DecHetero represents the equivalent of FL in decentralized settings with a lack of coordination between the devices regarding the models' initialization. Comparing its performance with standard FL, we observe that the additional source of heterogeneity is particularly detrimental and causes an additional degradation in the overall performance. Depending on the specific dataset, the devices can lose up to 3.39 percentage points in accuracy (MNIST) with respect to Federated Learning. The reason for such degradation is rooted in the averaging mechanism that, when applied to models initialized differently and trained on non-IID data, destroys the information previously learnt by the models.
Nonetheless, the averaging also acts as a (partial) synchronization event, after which the nodes find a partial agreement w.r.t. their neighbourhood. During the following updates, the divergence between models gets smaller and smaller, and the primary source of heterogeneity is due to the relative difference of their local datasets, from which FedAvg can recover, as suggested by the closeness of the final performance to the FL ones. 
Our solution does not suffer such shortcomings, although we still use the neighbourhood-wise average model to drive the devices' local model updates. Our difference-based update combined with the virtual teacher-based local training makes the models' updates less susceptible to both sources of heterogeneity, i.e., models initialization and non-IID-ness, by smoothing the models' parameters convergence in the neighbourhood and thus making the whole process less disruptive through time.
\subsubsection{Comparison against benchmark approaches}let us compare the performance with the decentralized direct competitors CFA and CFA-GE. The aggregation step of CFA is based on the weighted difference between the local and the neighbourhood models. The update is scaled proportionally to the size of the neighbourhood. In our case, the update is based on the difference between the local and neighbourhood-wide aggregate models. Such a slight variation significantly impacts the robustness of our approach to heterogeneity. Table \ref{tab:avg_acc_all} shows that Decdiff+VT outperforms CFA on all the datasets, improving the performance by 5.55 (MNIST), 3.33 (Fashion-MNIST), 8.46 (EMNIST) percentage points. CFA-GE aggregation step solves some of the weaknesses introduced by CFA by exploiting not only the information provided by the model parameters but also the gradients computed on the neighbourhood local data, as demonstrated by the average accuracy improvement reported in \autoref{tab:avg_acc_all}. It is worth noting that this improvement comes at an additional cost in terms of communication and computation, i.e., nodes exchange not only models but also gradients computed using previous versions of the neighbours' models, doubling the information transmitted in the system and increasing the computational effort by a factor that grows linearly with the size of the neighbourhood. Without additional communication or computational effort, Decdiff+VT show better or statistically equivalent performance than CFA-GE. 

\subsection{Ablation analysis}
\begin{table*}[ht]
 \centering
 \caption{Ablation analysis. \label{tab:ablation}}
\begin{tabular}{l|ll|ll|lll}
\toprule
\multirow{2}{*}{Method} & \multicolumn{2}{c |}{Loss} & \multicolumn{2}{c|}{Aggr. policy} & \multicolumn{3}{c}{Average Accuracy (Gain [\%pt])} \\
\cmidrule{2-8}
 & CE & VT & DecAvg & DecDiff & MNIST & Fashion & EMNIST \\
\midrule
DecHetero & x & & x & & 0.9071 & 0.8837 & 0.8301 \\
DecDiff & x &  &  & x & 0.9373 (+3.02\%pt) & 0.8838 (+0.01\%pt) & 0.8214 (-0.87\%pt) \\
DecDiff+VT &  & x &  & x & 0.9530 (+4.59\%pt) & 0.8904 (+0.67\%pt) & 0.8653 (+3.52\%pt) \\
\bottomrule
\end{tabular}
 \end{table*}
Let us analyze the contribution provided (i) by our aggregation policy DecDiff and (ii) by the loss function, including the virtual teacher, in the presence of data and model-initialization heterogeneity. To this end, we take DecHetero as a baseline. \autoref{tab:ablation} reports the relative performance of our approach, separating the contribution of the aggregation step from the one of the Virtual Teacher, compared to DecAvg. The difference between DecHetero and DecDiff regards only the aggregation function, since both use the cross-entropy loss. As we can see, our proposal in two out of three datasets is sufficient for improving the average performance of nodes in the system on simple datasets (MNIST and Fashion). However, on a more challenging task (EMNIST), it is insufficient to improve over the baseline. Substituting the traditional cross-entropy loss function on hard labels with the Virtual Teacher provides an additional improvement. Our solution DecDiff+VT improves by +1.57, +0.66, +4.39 percentage points on MNIST, Fashion and EMNIST, over DecDiff, respectively.

\subsection{Test loss analysis and characteristic time}

\begin{table}[t!]
\centering
\caption{Average characteristic time (in number of communication rounds) w.r.t. the accuracy of the centralised benchmark.}
\label{tab:conv_time}
\begin{tabular}{llrrrr}
  \toprule
Dataset & Method & 50\% & 80\% & 90\% & 95\% \\ 
  \midrule
\multirow{7}{*}{MNIST}& DecHetero & 97.40 & 235.60 & 583.25 & -  \\ 
  &CFA & 109.40 & 322.40 & 609.67 & - \\ 
  &CFA-GE & 102.80 & 173.40 & \bf{327.60} & 691.60 \\ 
  &DecDiff & \bf{52.80} & \bf{145.40} & 395.60 & 671.00 \\ 
  &DecDiff+VT & 55.20 & 161.00 & 360.00 & \bf{540.75} \\ 
  \cmidrule{2-6}
  &FED & 11.80 & 57.40 & 307.40 & 662.67 \\ 
  &ISOL & 262.20 & - & - & - \\ 
   \midrule
\multirow{7}{*}{Fashion}&DecHetero & \bf{3.20} & \bf{53.20} & 387.20 & - \\ 
  & CFA & 10.60 & 142.40 & - & - \\ 
  & CFA-GE & 10.20 & 118.40 & 693.60 & -\\ 
  & DecDiff & 6.80 & 54.00 & 319.80 & -\\ 
  &DecDiff+VT & 7.00 & 53.80 & \bf{313.60} & -\\ 
  \cmidrule{2-6}
  &FED & 1.20 & 25.00 & 216.00 & -\\ 
  &ISOL & 3.80 & - & - & -\\ 
   \midrule
\multirow{7}{*}{EMNIST}&DecHetero & 22.00 & \bf{91.00} & 370.50 & - \\ 
  &CFA & 62.80 & 418.00 & - & - \\ 
  &CFA-GE & 33.20 & 167.00 & 324.20 & 812.25 \\ 
  &DecDiff & 17.60 & 107.20 & 566.75 & - \\ 
  &DecDiff+VT & \bf{9.60} & 106.80 & \bf{138.40} & \bf{400.25} \\ 
  \cmidrule{2-6}
  &FED & 4.00 & 53.80 & 322.25 & - \\ 
  &ISOL & 28.00 & - & - & - \\ 
  \bottomrule
\end{tabular}
\end{table}

\autoref{tab:avg_acc_all} shows that our Decdiff+VT solution consistently outperforms the competitors in terms of accuracy. Let us now analyze how the learning evolves over the communications rounds. To this end, refer to \autoref{tab:conv_time} where we report the characteristic times w.r.t the accuracy of the centralized benchmark for all the considered methods, including FED and ISOL, even though they are not direct competitors. We highlight the minimum time for each relative accuracy, i.e., the fastest method that reaches a certain relative accuracy w.r.t. the centralized benchmark. Let us focus on the decentralized methods.
Our solution, DecDiff+VT, is the fastest on all the datasets. Precisely, on MNIST and EMNIST, only DecDiff+VT and CFA-GE consistently exceed the 95\% threshold, and the former is 1.28$\times$ and 2$\times$ faster than the latter, respectively. In Fashion-MNIST, although none of the considered solutions reach the 95\% threshold but only the 90\%, DecDiff+VT is 2.2$\times$ faster than CFA-GE. 

\begin{figure}
     \centering
     \subfloat[]{
         \includegraphics[width=1.0\columnwidth]{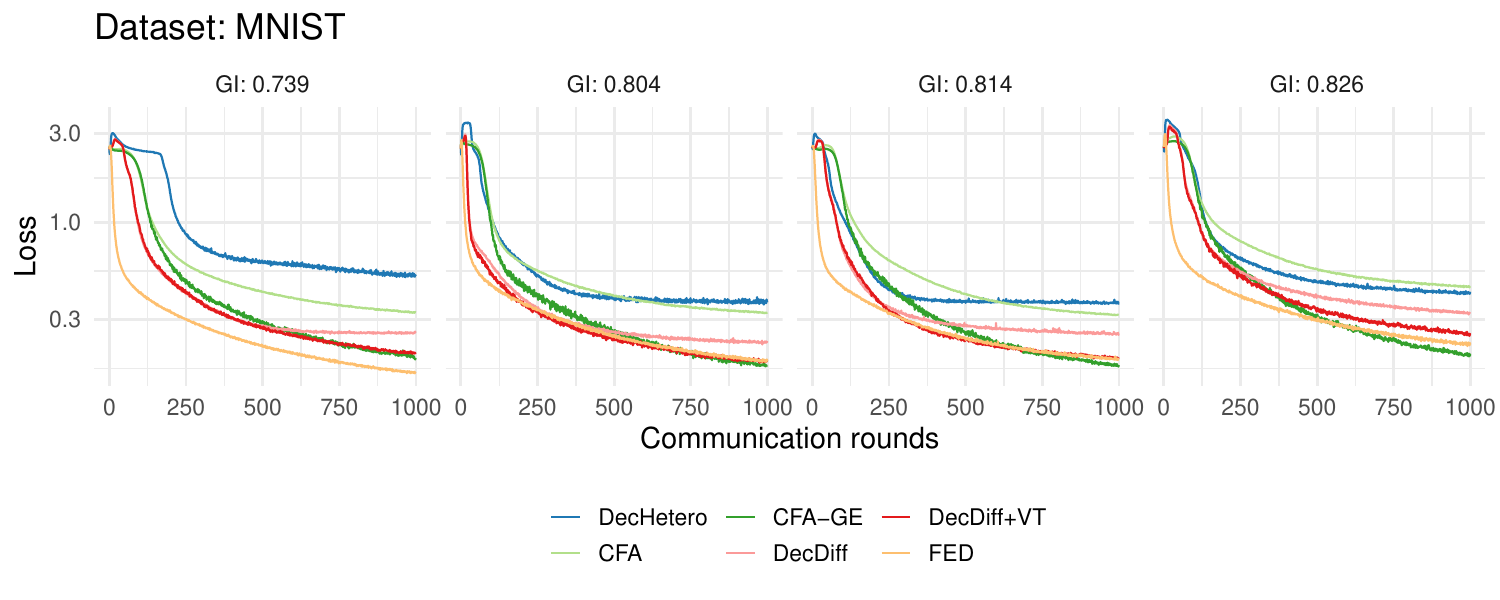}
         \label{fig:mnist_test}}
     \hfill
     \subfloat[]{
         \includegraphics[width=1\columnwidth]{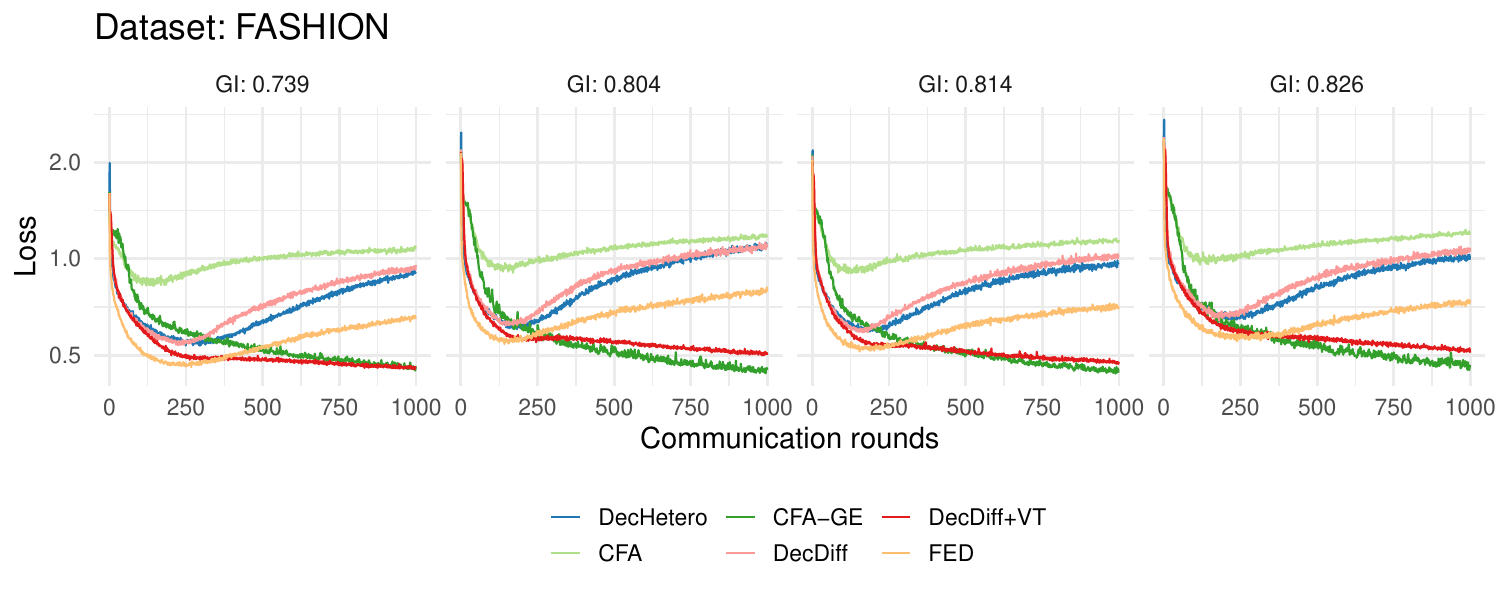}
         \label{fig:fashion_test}
    }
     \hfill
    \subfloat[]{
         \includegraphics[width=1\columnwidth]{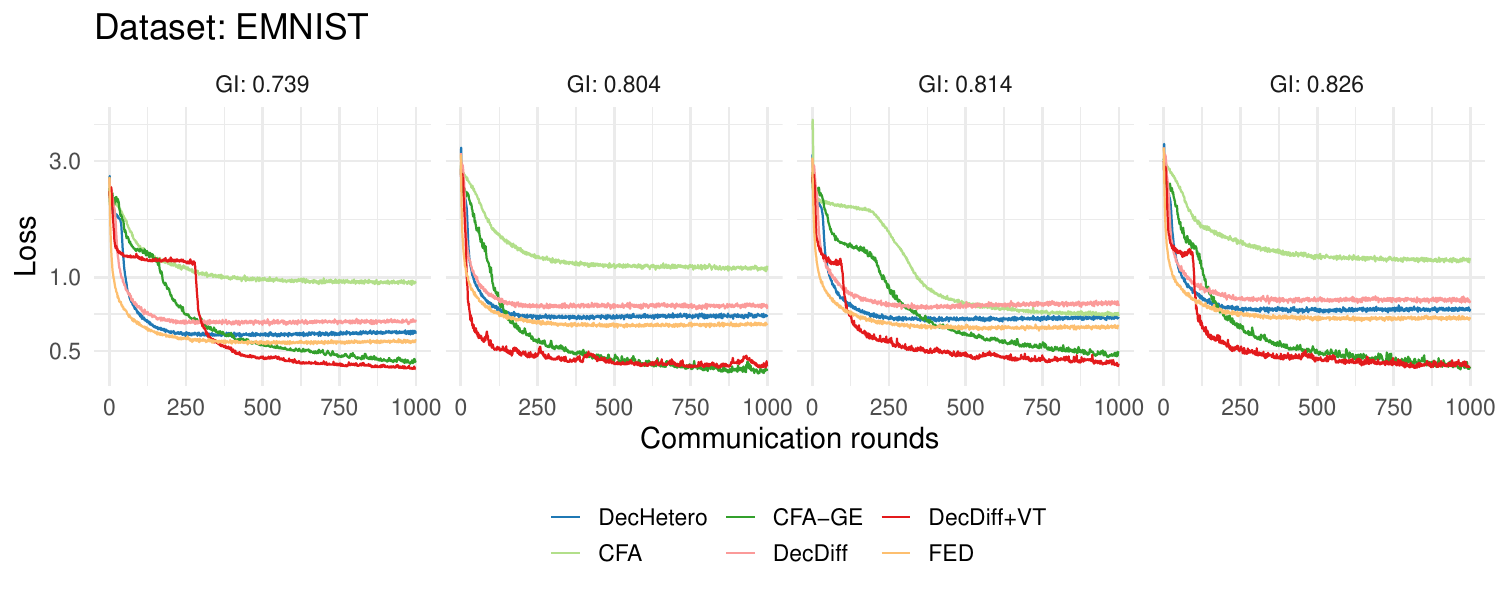}
         \label{fig:emnist_test}}
        \caption{Test loss curves for each dataset and different levels of non-IID-ness in the data distribution, quantified by the Gini Index (GI). DecDiff+VT and CFA-GE are the only approaches preventing overfitting across device models. }
        \label{fig:test_acc}
\end{figure}

Interestingly, DecDiff and DecDiff+VT are generally faster than (MNIST and EMNIST) or comparable to (Fashion) competitors in embedding constructively the knowledge contained in the received models during the initial communication rounds. Moreover, DecDiff+VT shows the ability to constantly learn and improve even when the other competitors stop learning and start to overfit the local data. As shown in Figures \ref{fig:mnist_test},\ref{fig:fashion_test},\ref{fig:emnist_test}, it is clear that for CFA-GE and DecDiff-VT the test loss keeps its descent when the other ones start slowing down or even increasing. These two methods show very similar behaviour, which is impressive if we consider that DecDiff+VT does not exploit any of the additional information contained in the gradients received from the neighbourhood as in CFA-GE. Moreover, apart from the initial communication rounds, DecDiff+VT is generally faster than CFA-GE.

% \begin{figure}[t]
% \centering
% \includegraphics[width=1.0\columnwidth]{figures/loss_over_commrounds_mean_test-mnist.pdf}
% \label{fig:mnist_test}
% \end{figure}
% \begin{figure}[t]
% \centering
% \includegraphics[width=1\columnwidth]{figures/loss_over_commrounds_mean_test-fashion.pdf}
% \label{fig:fashion_test}
% \end{figure}
% \begin{figure}[t]
% \centering
% \includegraphics[width=1\columnwidth]{figures/loss_over_commrounds_mean_test-emnist.pdf}
% \label{fig:emnist_test}
% \end{figure}

\subsection{Node-wise analysis}
\begin{figure}
     \centering
     \subfloat[]{
         \includegraphics[width=1.0\columnwidth]{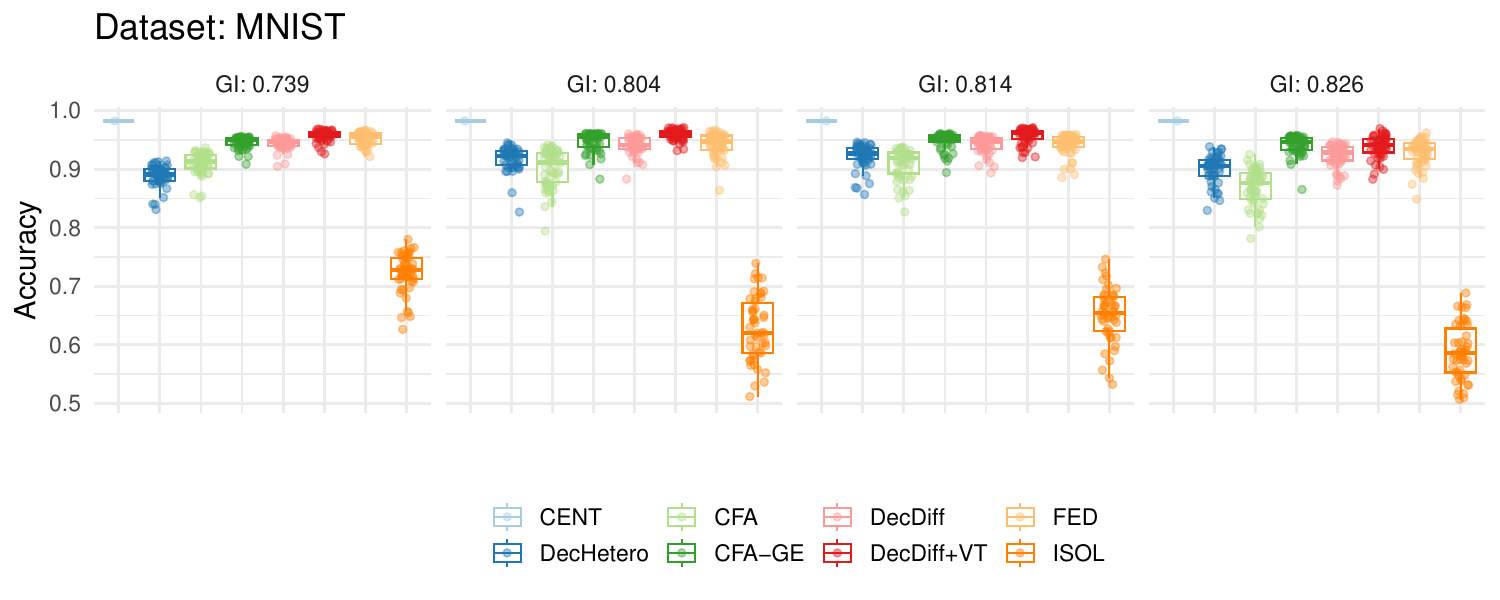}
         \label{fig:mnist_box}}
     \hfill
    \subfloat[]{
         \includegraphics[width=1\columnwidth]{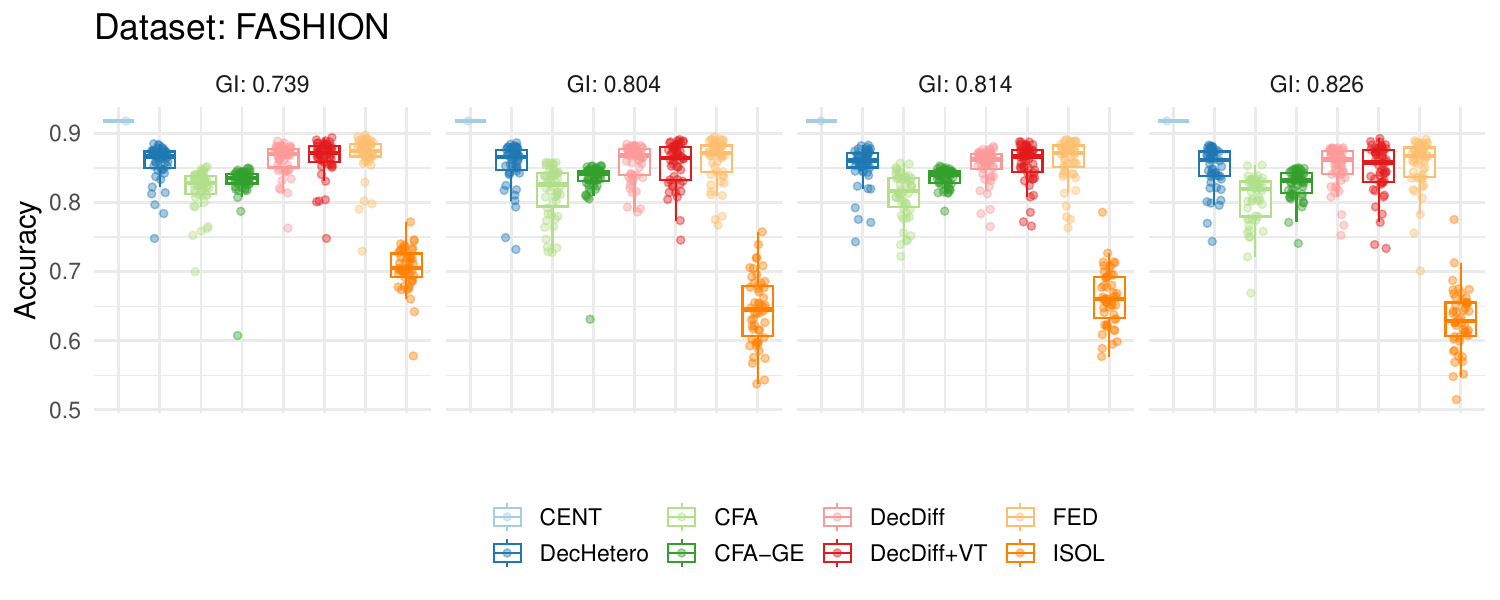}
         \label{fig:fashion_box}}
     \hfill
     \subfloat[]{
         \includegraphics[width=1\columnwidth]{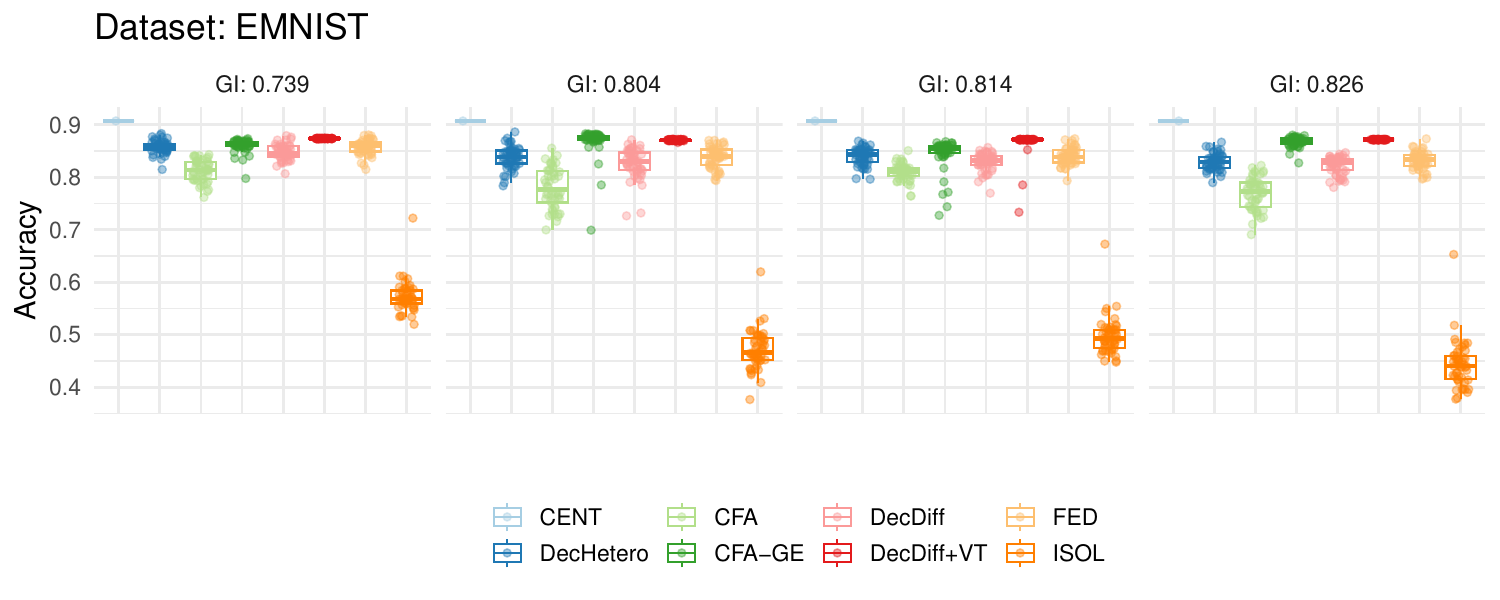}
         \label{fig:emnist_box}}
        \caption{Boxplot of node-wise accuracy at the last communication round for each dataset and different levels of non-IID-ness in the data distribution, quantified by the Gini Index (GI). DecDiff+VT and CFA-GE result in a more concentrated distribution of accuracy.}
        \label{fig:pernode_box}
\end{figure}

Until now, we analyzed the aggregate performance of all the nodes in the system, which reflects the system's global behaviour. In the following, we analyze the performance of the single nodes at the last communication round. Figure~\ref{fig:pernode_box} reports, for each methodology and each dataset, the accuracy achieved by each node in the system. The first interesting aspect to observe regards the dispersion of accuracy that the nodes achieve in isolation (ISOL), particularly evident in Fashion-MNIST and MNIST. The effect of cooperation is evident by looking at the accuracy distribution of the decentralized approaches. However, for some of them (i.e., DecHetero, CFA, DecDiff), the accuracy distribution still appears rather dispersed, suggesting that, in highly heterogeneous settings, neither a na\"ive application of the decentralized version of FedAvg nor some more sophisticated approaches (i.e., CFA, DecDiff) are able to make the nodes exploit all the knowledge contained in their local data. Differently, only DecDiff+VT and CFA-GE result in a more concentrated distribution of accuracy, highlighting in our solution the role of the Virtual Teacher, as it becomes evident in Figure~\ref{fig:emnist_box}. 

%=========================================================
\ifextremehetero
\subsection{Performance under extreme data heterogeneity}
\label{sec:extreme_hetero}

TBD

\fi
%=========================================================
% \subsection{Sensitivity analysis of $\beta$ parameter}
% \label{sec:sensitivity}

%=========================================================
%!TEX root = ./tnnls_main.tex
\section{Conclusion}
\label{sec:conclusion}
In this paper, we consider the problem of performing a decentralised, federated learning task in a highly pervasive and uncoordinated environment. We assume that the devices involved, beyond holding non-IID data, are also interconnected through a complex network graph, and their interactions are limited to their neighbourhood only. Due to the lack of coordination, the devices in the system do not agree on the same model's parameter configuration, resulting in a learning problem with a high degree of heterogeneity (both from data and models). We propose a Decentralised Learning scheme for overcoming these difficulties that factor in two blocks: an aggregation function that updates the potentially feature-wise incompatible models without destroying previous information and a local training procedure based on knowledge distillation and soft-labelling for improving the generalisation of the local models. 
In terms of accuracy our approach evidently outperforms
% Our approach shows superior accuracy performance over
standard benchmarks, avoiding or mitigating overfitting without paying extra costs in communication during the process. 
While our primary focus in this study has centered on addressing the lack of coordination within a fully decentralized setting characterized by heterogeneous data and model initialization, we aim to expand our research by considering further sources of heterogeneity. These may include variations in model architecture, such as nodes employing a mix of MLPs and CNNs, discrepancies in model training approaches, such as nodes utilizing different numbers of training epochs for their local training procedures, or different and constrained computational resources.

\section*{Acknowledgments}
This work was partially supported by the H2020 HumaneAI Net (\#952026), H2020 INFRAIA-01-2018-2019 SoBigData++ (\#871042),  and by the CHIST-ERA-19-XAI010 SAI projects, FWF (grant No. I 5205). C. Boldrini's work was partly funded by the PNRR - M4C2 - Investimento 1.3, Partenariato Esteso PE00000013 - "FAIR", A. Passarella's work was partly funded by the PNRR - M4C2 - Investimento 1.3, Partenariato Esteso PE00000001 - "RESTART", both funded by the European Commission under the NextGeneration EU programme. J. Kertész and G. Iñiguez acknowledge support from AFOSR (Grant No. FA8655-20-1-7020). J. Kertész, G. Iñiguez and M. Karsai acknowledge the constructive insights of Pablo Lozano and Arash Badie Modiri.

% {\appendix[Proof of the Zonklar Equations]
% Use $\backslash${\tt{appendix}} if you have a single appendix:
% Do not use $\backslash${\tt{section}} anymore after $\backslash${\tt{appendix}}, only $\backslash${\tt{section*}}.
% If you have multiple appendixes use $\backslash${\tt{appendices}} then use $\backslash${\tt{section}} to start each appendix.
% You must declare a $\backslash${\tt{section}} before using any $\backslash${\tt{subsection}} or using $\backslash${\tt{label}} ($\backslash${\tt{appendices}} by itself
%  starts a section numbered zero.)}

%{\appendices
%\section*{Proof of the First Zonklar Equation}
%Appendix one text goes here.
% You can choose not to have a title for an appendix if you want by leaving the argument blank
%\section*{Proof of the Second Zonklar Equation}
%Appendix two text goes here.}

\balance

\bibliographystyle{IEEEtran}
\bibliography{references_v1}

% \newpage

% \section{Biography Section}
% If you have an EPS/PDF photo (graphicx package needed), extra braces are
%  needed around the contents of the optional argument to biography to prevent
%  the LaTeX parser from getting confused when it sees the complicated
%  $\backslash${\tt{includegraphics}} command within an optional argument. (You can create
%  your own custom macro containing the $\backslash${\tt{includegraphics}} command to make things
%  simpler here.)
 
% \vspace{11pt}

% \bf{If you include a photo:}\vspace{-33pt}
% \begin{IEEEbiography}[{\includegraphics[width=1in,height=1.25in,clip,keepaspectratio]{fig1}}]{Michael Shell}
% Use $\backslash${\tt{begin\{IEEEbiography\}}} and then for the 1st argument use $\backslash${\tt{includegraphics}} to declare and link the author photo.
% Use the author name as the 3rd argument followed by the biography text.
% \end{IEEEbiography}

% \vspace{11pt}

% \bf{If you will not include a photo:}\vspace{-33pt}
% \begin{IEEEbiographynophoto}{John Doe}
% Use $\backslash${\tt{begin\{IEEEbiographynophoto\}}} and the author name as the argument followed by the biography text.
% \end{IEEEbiographynophoto}
% \vfill

\end{document}